\documentclass{article}

\usepackage{arxiv}

\usepackage[utf8]{inputenc} 
\usepackage[T1]{fontenc}    
\usepackage{hyperref}       
\usepackage{url}            
\usepackage{booktabs}       
\usepackage{amsfonts}       
\usepackage{nicefrac}       
\usepackage{microtype}      
\usepackage{lipsum}		
\usepackage{graphicx}
\usepackage{cite}
\usepackage{doi}

\title{A Category-Fragment Segmentation Framework for Pelvic Fracture Segmentation in X-ray Images}

\author{
\href{https://orcid.org/0009-0001-0905-3833}{\includegraphics[scale=0.06]{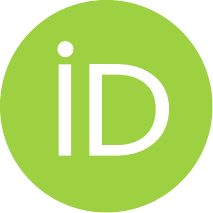}\hspace{1mm}Daiqi Liu}\thanks{Corresponding author: daiqi.deutschfau.liu@fau.de} \quad
\href{https://orcid.org/0000-0001-9602-6237}{\includegraphics[scale=0.06]{orcid.pdf}\hspace{1mm}Fuxin Fan} \quad
\href{https://orcid.org/0000-0000-0000-0000}{\includegraphics[scale=0.06]{orcid.pdf}\hspace{1mm}Andreas Maier} \\
Pattern Recognition Lab, Friedrich-Alexander-Universität Erlangen-Nürnberg \\
\texttt{\ daiqi.deutschfau.liu@fau.de, fuxin.fan@fau.de, andreas.maier@fau.de}
}

\date{}

\renewcommand{\headeright}{Liu, Fan \& Maier}
\renewcommand{\undertitle}{}
\renewcommand{\shorttitle}{Pelvic Fracture Segmentation Network}

\hypersetup{
pdftitle={A Category-Fragment Segmentation Framework for Pelvic Fracture Segmentation in X-ray Images},
pdfauthor={Daiqi Liu, Fuxin Fan, Andreas Maier},
}

\begin{document}
\maketitle

\begin{abstract}
Pelvic fractures, often caused by high-impact trauma, frequently require surgical intervention. Imaging techniques such as CT and 2D X-ray imaging are used to transfer the surgical plan to the operating room through image registration, enabling quick intraoperative adjustments. Specifically, segmenting pelvic fractures from 2D X-ray imaging can assist in accurately positioning bone fragments and guiding the placement of screws or metal plates. In this study, we propose a novel deep learning-based category and fragment segmentation (CFS) framework for the automatic segmentation of pelvic bone fragments in 2D X-ray images. This framework consists of three consecutive steps. First, the category segmentation network extracts the left and right ilia and sacrum from X-ray images. Then, the fragment segmentation network further isolates the fragments in each masked bone region. Finally, the initially predicted bone fragments are reordered and refined through post-processing operations to form the final prediction. In the best-performing model, segmentation of pelvic fracture fragments achieves an Intersection over Union (IoU) of 0.91 for anatomical structures and 0.78 for fracture segmentation. Experimental results demonstrate that our CFS framework is effective in segmenting pelvic categories and fragments. The source code are publicly available at \url{https://github.com/DaE-plz/CFSSegNet} for further research and development.
    
\end{abstract}

\section{Introduction}
Pelvic fractures, often caused by high-energy impacts like motor vehicle accidents or  falls, can result in mortality rates exceeding 50\% in severe cases \cite{3639-11,3639-12}. Accurate segmentation of pelvic fractures in 3D scans and X-ray images is essential, as it enables precise localization of fracture sites and guides both manual and automated reduction planning to restore pelvic anatomy \cite{3639-20}. While 3D CT scans provide detailed anatomical insights and are widely used for preoperative planning, 2D X-rays images are vital for intraoperative guidance and surgical plan execution. However, the projection nature of X-ray imaging complicates the automatic segmentation of pelvic bone fragments due to overlapping anatomy, fracture displacement, and complex pelvic morphology.

Most existing methods for pelvic fracture segmentation focus on CT data due to its volumetric nature and high contrast resolution \cite{3639-22,3639-23} . Deep learning approaches have been successfully applied to segment pelvic bones and detect fractures in CT scans, leveraging the rich spatial information available in 3D data \cite{3639-01}. In contrast, there is a lack of research on the automatic segmentation of pelvic fractures in X-ray images. Traditional image processing techniques, such as edge detection and thresholding, struggle with the complexity and variability of pelvic anatomy in 2D projections \cite{3639-15,3639-16}. Some studies have explored convolutional neural networks (CNNs) for bone segmentation in X-ray images \cite{3639-02}. Additionally, recent studies have utilized deep learning for fracture detection in X-ray images, but they typically do not provide detailed segmentation of individual bone fragments \cite{3639-17,3639-18,3639-19}.

This paper presents a deep learning-based method to automatically segment pelvic fracture categories and fragments from preoperative X-ray images. Our major contributions are as follows: (1) Our pipeline is proposed a category and fragment segmentation pipeline for pelvic fracture segmentation on X-ray images that achieves high accuracy in segmenting bone fragments. (2) During the data preparation stage, zero-padding is applied to both images and masks. (3) As a post-processing step, the predicted bone fragment masks are multiplied by the category masks to refine the final output.

\section{Materials and methods}

\begin{figure}[b]
  \centering
  \includegraphics[width=0.8\linewidth]{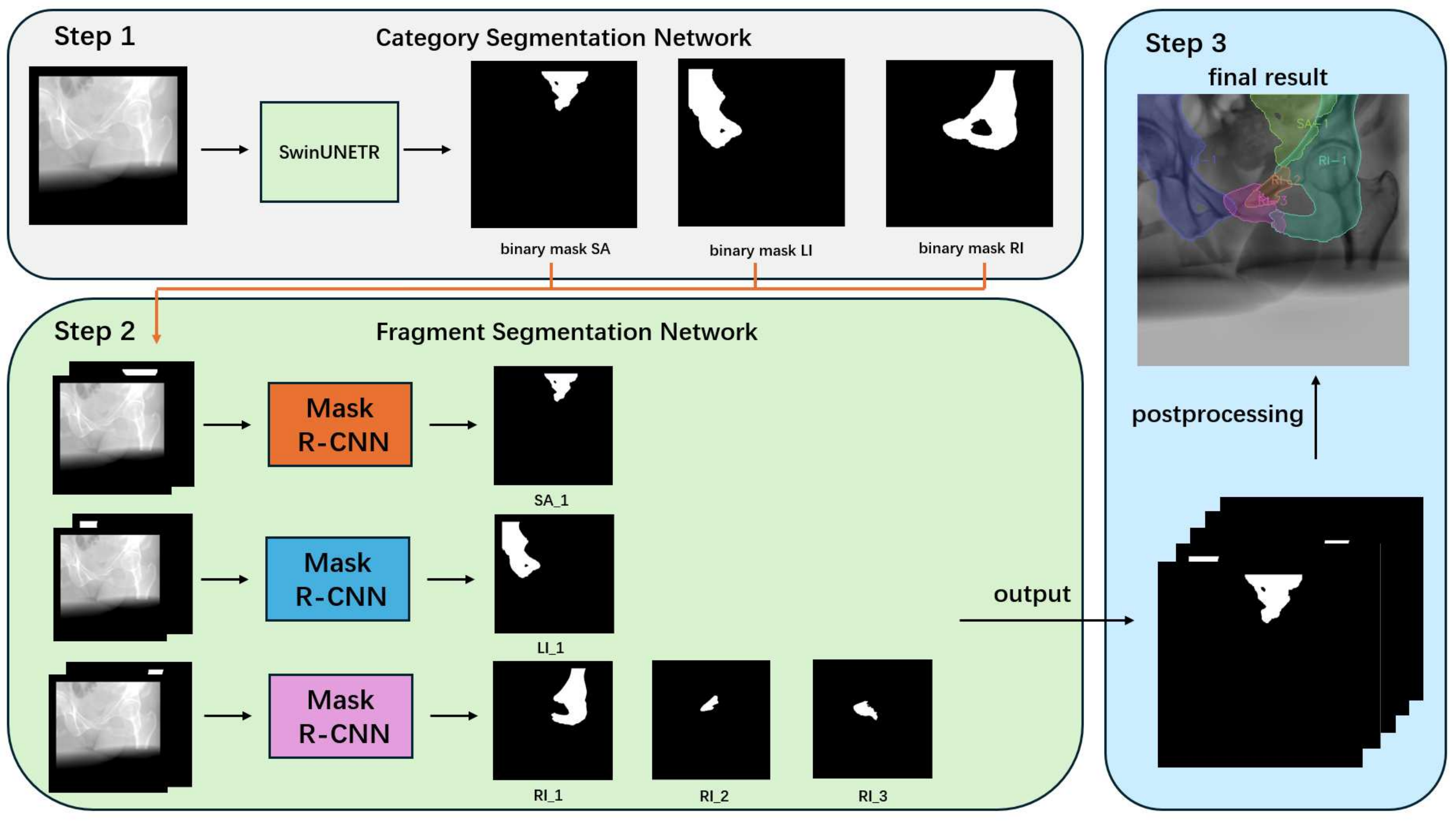}
  \caption{Overview of the proposed pelvic fracture segmentation framework. The framework is divided into three main components: category segmentation, fragment segmentation, and post-processing.}
  \label{3639-framework}
\end{figure}

Our study aims to automatically segment the fragments in different categories of pelvic bones (the left and right ilia and the sacrum) from X-ray images.  As illustrated in Figure.~\ref{3639-framework}, the whole CFS framework consists of three steps.

The first step aims to segment each part of the pelvic bones from X-ray images. We utilize three separate category segmentation networks, each based on the Swin UNETR \cite{3639-08} architecture, to extract the left ilium, right ilium, and sacrum. The second step tries to isolate the bone fragments within each segmented bone region. We employ three separate fragment segmentation networks, each based on the Mask R-CNN \cite{3639-9} architecture, which not only performs segmentation but also integrates object detection through bounding box prediction and classification. These networks take both the X-ray images and the binary masks generated in the first step as input channels, allowing the model to focus on specific bone regions. During inference, the network predicts multiple bounding boxes, each associated with a confidence score. Only boxes with a confidence score above a predefined threshold, which is 0.8 in this work, are selected as potential bone fragments. In the third step, post-processing techniques are applied. The category masks from the first step are multiplied by the fragment segmentation masks from the second step. Additionally, the predicted masks are reordered according to predefined indices to ensure consistency in the output.

The dataset used in this work is sourced from the MICCAI challenge, PENGWIN \cite{3639-21}, which provides both 2D X-ray images and 3D CT scans with annotated labels for pelvic category and fragment segmentation. The 2D X-ray images are generated from 3D CT data using DeepDRR \cite{3639-10}, a tool that produces highly realistic images by accurately simulating physical properties and clinical detail, including anatomical features, intensity, and noise patterns. The final dataset consists of 50,000 X-ray images generated from 100 CT volumes, with each CT volume producing 500 images from different angles. The input images and masks have a resolution of $448 \times 448$ pixels. For each X-ray image, the dataset provides binary masks for all bone fragments within each pelvic category, with the number of fragments varying across cases.

\begin{table}[t]
\centering
\caption{Segmentation results of different architectures used for category segmentation in the CFS framework. Swin UNETR +p represents the effect of adding zero padding operation on the experiment. The evaluation metrics are presented as mean with standard deviation (SD) for IoU, ASSD, and HD95 (on a test set of 5,000 cases). -C indicates the result of category segmentation, -F indicates the result of fragment segmentation.}

\begin{tabular*}{\textwidth}{l@{\extracolsep\fill}llllll}
\hline
Model   & IOU-C  & ASSD-C  & HD95-C  & IOU-F  & ASSD-F  & HD95-F  \\
\hline
nnU-Net         & 0.890  & 4.46   & 15.24  & 0.734 & 11.14   & 43.64 \\
                &  (0.07)  &  (0.79)  &  (3.02)  & (0.16) & (3.08)  &  (4.35) \\
                \hline
UNet++          & 0.896   & 4.45  & 15.23   & 0.736   & 10.55   & 41.49  \\
        &  (0.07)  & (0.72)   &  (2.49)  &  (0.15)  &  (2.54)  & (3.98) \\
        \hline
Attention U-Net & 0.898   & 3.25  & 14.78   & 0.745   & 10.35  & 40.87  \\
 & (0.06)  &  (0.70)  &  (2.61)  &  (0.15)  &  (2.69) &  (4.04) \\
 \hline
VM-UNet         & 0.871    & 5.76   & 17.85 & 0.724  & 11.36   & 42.99 \\
         &  (0.09)   & (1.14)  &  (3.52)  &  (0.17) &  (3.54)  &  (4.63) \\
         \hline
Swin UNETR      & 0.901   & 3.02   & 14.56   & 0.753   & 9.83   & 40.25  \\
    &  (0.07)    & (0.70)  &  (2.26)  & (0.11)  & (2.35)  &(3.74) \\
    \hline
Swin UNETR +p   & 0.914     & 2.25   & 13.28   & 0.775   & 9.53  & 38.65  \\
  &  (0.06)    & (0.49)   &  (2.21)  &  (0.10)  & (2.28)   & (3.44) \\
\hline
\end{tabular*}
\label{3639-result_category_segmentation}
\end{table}

All models were implemented using PyTorch 2.1.2. The experiments were performed on an Nvidia A100 GPU. The comprehensive code are provided at our GitHub repository: \url{https://github.com/DaE-plz/CFSSegNet}.

\section{Results}

Table.~\ref{3639-result_category_segmentation} presents the performance comparison of various neural network architectures used for category segmentation in Step 1 of the CFS framework on a test set of 5,000 images. The architectures tested include nnU-Net \cite{3639-03,3639-04}, UNet++ \cite{3639-05}, Attention U-Net \cite{3639-06}, VM-UNet \cite{3639-07} and Swin UNETR \cite{3639-08}. Additionally, we conducted experiments to assess the impact of using zero-padding within the CFS framework, specifically with the Swin UNETR architecture (Swin UNETR +p).

For category segmentation, the results show that nnU-Net and UNet++ achieve comparable performance, with both models exhibiting a similar Intersection over Union (IoU) score of approximately 0.89. Attention U-Net demonstrates a slightly better performance in category segmentation, with an IoU of 0.898 and lower ASSD and HD95 values. Swin UNETR further improves the category segmentation results, delivering an IoU of 0.901. The Swin UNETR +p model performs the best among all architectures, achieving the highest IoU (0.914) with the SD of 0.06 and the lowest ASSD and HD95 scores, making it the most effective for category segmentation.

For fragment segmentation, overall performance is generally lower compared to category segmentation, as seen in the lower IoU and higher ASSD and HD95 values. However, the models follow a similar trend as in category segmentation. Swin UNETR +p again demonstrates the best performance, achieving an IoU of 0.775 with the SD of 0.10 and providing more accurate fragment segmentation, as indicated by the lowest ASSD (9.53) and HD95 (38.65) values.

\begin{figure}[h]
    \centering
    \includegraphics[width=0.8\linewidth]{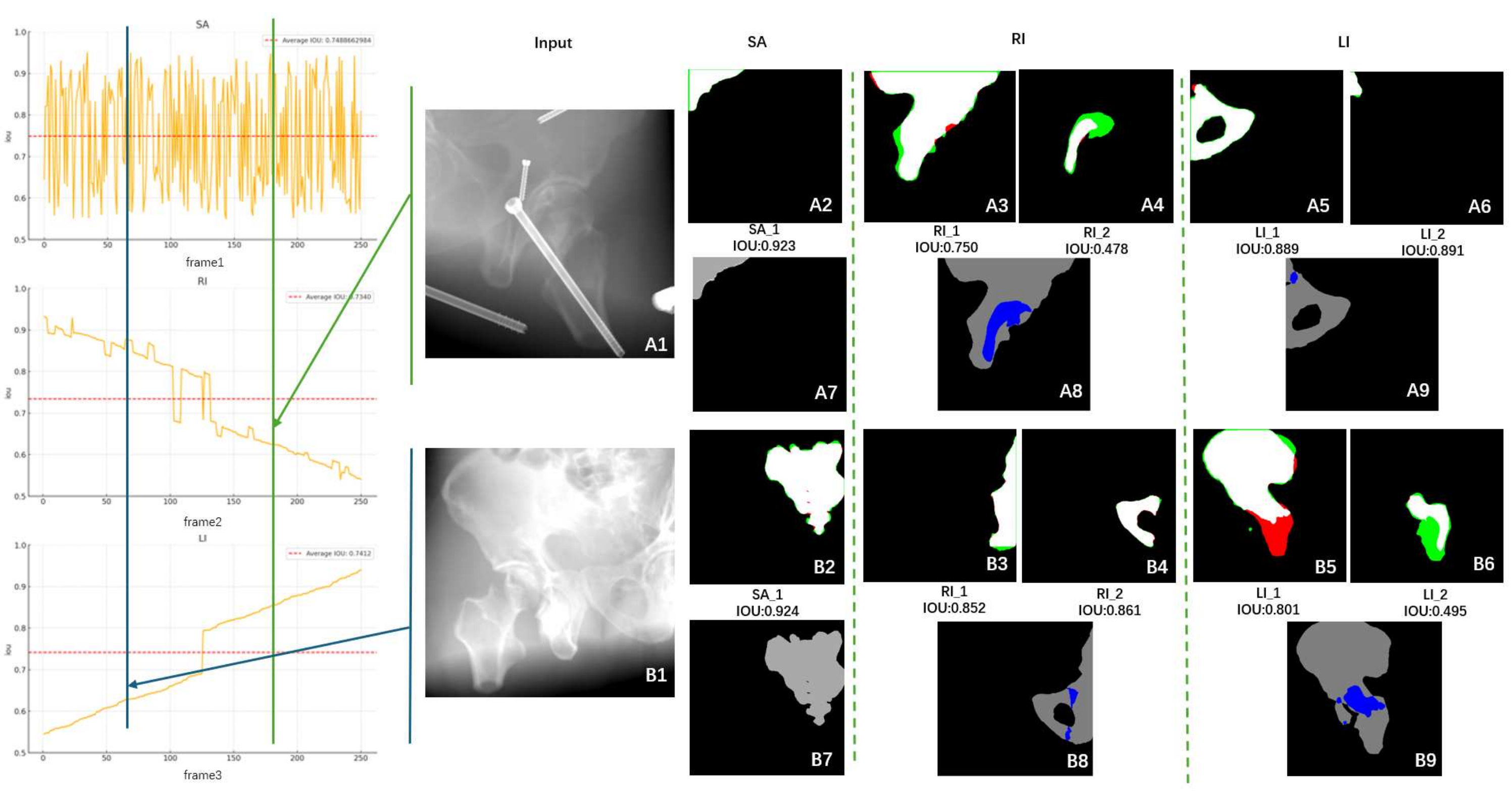}
    \caption{Relationship between anatomical overlap and segmentation performance. blue represents the overlapping regions, red indicates false positives, and green indicates false negatives.}
    \label{3639-overlap}
\end{figure}

We used 250 X-ray images generated from a single CT volume. With the left hipbone as a reference, the overlap ratio (the proportion of the left hipbone area that is overlapped) was calculated for each image. The images were subsequently sorted by overlap ratio, from highest to lowest. For each sorted image, IoU scores were computed for fragment segmentation of the sacrum, right hipbone, and left hipbone. The left side of Figure.~\ref{3639-overlap} presents the IoU scores for bone fragment segmentation corresponding to each X-ray image, with the X-axis representing the image index and the Y-axis representing the IoU score for each bone fragment. On the right side of Figure.~\ref{3639-overlap}, two representative examples are shown, labeled A1 to A9 (top) and B1 to B9 (bottom). Each example consists of two rows: the upper row displays the predicted segmentation masks, where green areas indicate false negatives and red areas represent false positives. The lower row highlights the overlapping regions between bone fragments, shown in blue. 

In the example labeled A, image A8 shows a large overlapping area at RI, with an IoU score of 0.478 for the bone fragment RI\textsubscript{2}, while image A9 shows minimal overlap at LI, achieving an IoU score of 0.889 for the fragment LI\textsubscript{1}. In the example labeled B, image B8 has minimal overlap in the RI region with an IoU score of 0.852 for RI\textsubscript{1}, whereas image B9 shows large overlap in the LI region with an IoU score of 0.495 for LI\textsubscript{2}.

\section{Discussion}

The results from Table.~\ref{3639-result_category_segmentation} demonstrate that models incorporating more advanced architectures, such as attention mechanisms (Attention u-net) or transformer-based structures (Swin UNETR), enhance segmentation performance, particularly in capturing boundary details. However, the complexity of a model does not necessarily correlate with improved segmentation performance. For instance, VM-UNet, despite its complexity, did not perform well for pelvic bone segmentation. When examining the effect of zero-padding on segmentation performance, we observed that using padded images and masks for training improved boundary segmentation accuracy and led to better overall performance. Additionally, using our best model for testing, the SD of IoU in fragment segmentation (0.10) was higher than that in category segmentation (0.06), indicating bigger inconsistency. This larger SD suggests that the model is less stable when segmenting bone fragments compared to category segmentation.

From Figure.~\ref{3639-overlap}, the results indicate that images with higher overlap ratios tend lower IoU scores for fragment segmentation. This analysis underscores the challenge that anatomical overlap poses for bone fragment segmentation, increased overlap in X-ray projections obscures the boundaries between bone fragments, making it more difficult for the network to achieve accurate segmentation. For future work, we plan to explore whether pre-segmenting overlapping regions can improve the overall performance of bone fragment segmentation.

\section*{Acknowledgements}
The authors gratefully acknowledge the scientific support and HPC resources provided by the Erlangen National High Performance Computing Center (NHR@FAU) of the Friedrich-Alexander-Universität Erlangen-Nürnberg (FAU). The hardware is funded by the German Research Foundation (DFG).

\bibliographystyle{unsrt}







\end{document}